\documentclass[10pt,twocolumn,letterpaper]{article}

\usepackage{cvpr}
\usepackage{times}
\usepackage{epsfig}
\usepackage{graphicx}
\usepackage{amsmath}
\usepackage{amssymb}

\usepackage{bm}
\usepackage{times}
\usepackage{amsfonts}
\usepackage[ruled,vlined]{algorithm2e}
\usepackage{multirow} 
\usepackage{mathrsfs}
\usepackage{subfigure}
\usepackage{graphicx,color}
\usepackage{url}
\usepackage{amsthm}

\usepackage[breaklinks=true,bookmarks=false]{hyperref}
\usepackage[hyphenbreaks]{breakurl}

\cvprfinalcopy 

\def\cvprPaperID{****} 

\begin{document}

\title{Support Vector Guided Softmax Loss for Face Recognition}
\author{Xiaobo Wang$^{\rm{ 1}}$, Shuo Wang$^{\rm{ 1}}$, Shifeng Zhang$^{\rm{ 2}}$, Tianyu Fu$^{\rm{ 1}}$, Hailin Shi$^{\rm{1}}$, Tao Mei$^{\rm{ 1}}$\\
{$^{\rm 1}$JD AI Research
$^{\rm 2}$ Institute of Automation, Chinese Academy of Science. }\\
{\tt\small \{wangxiaobo8,wangshuo30,futianyu,shihailin,tmei\}@jd.com, shifeng.zhang@nlpr.ia.ac.cn}
}

\maketitle

\begin{abstract}
   Face recognition has witnessed significant progresses due to the advances of deep convolutional neural networks (CNNs), the central challenge of which, is feature discrimination. To address it, one group tries to exploit mining-based strategies (\textit{e.g.}, hard example mining and focal loss) to focus on the informative examples. The other group devotes to designing margin-based loss functions (\textit{e.g.}, angular, additive and additive angular margins) to increase the feature margin from the perspective of ground truth class. Both of them have been well-verified to learn discriminative features. However, they suffer from either the ambiguity of hard examples or the lack of discriminative power of other classes. In this paper, we design a novel loss function, namely support vector guided softmax loss (SV-Softmax), which adaptively emphasizes the mis-classified points (support vectors) to guide the discriminative features learning. So the developed SV-Softmax loss is able to eliminate the ambiguity of hard examples as well as absorb the discriminative power of other classes, and thus results in more discrimiantive features. To the best of our knowledge, this is the first attempt to inherit the advantages of mining-based and margin-based losses into one framework. Experimental results on several benchmarks have demonstrated the effectiveness of our approach over state-of-the-arts.
\end{abstract}

\section{Introduction}
Face recognition is a fundamental and of great practice values task in the community of computer vision and patter recognition. The task of face recognition contains two categories, face identification to classify a given face to a specific identity, and face verification to determine whether a pair of face images are of the same identity.
Though it has been extensively studied for decades \cite{wright2009robust,cai2014support,USSDL,DeepID,DeepID2+,shi2017cross}, there still exist a great many challenges for accurate face recognition, especially on large-scale test datasets, such as MegaFace Challenge \cite{megaface_1} or Trillion Pairs Challenge\footnote{\url{http://trillionpairs.deepglint.com/overview}}.

In recent years, the advanced face recognition models are usually built upon deep convolutional neural networks \cite{Attention56,Resnet,VGG} and the learned discriminative features play a significant role. To train deep models, the CNNs are generally equipped with classification loss functions \cite{Deepface,NormFace,Center,SM-Softmax,SphereFace,EM-Softmax}, metric learning loss functions \cite{Contrastive,Facenet,Angular} or both \cite{DeePFR,DeepID2+,Center,zheng2018ring}. Metric learning loss functions such as contrastive loss \cite{Contrastive} or triplet loss \cite{Facenet} usually suffer from high computational cost. To avoid this problem, they require carefully designed sample mining strategies and the performance is very sensitive to these strategies. So increasingly more researchers shift their attentions to construct deep face recognition models by re-designing the classification loss functions.

Intuitively, face features are discriminative if their intra-class compactness and inter-class separability are well maximized. However, as pointed out by many recent studies \cite{Center,NormFace,SphereFace,AM-Softmax,EM-Softmax,Arc-Softmax}, the current prevailing classification loss function (\textit{i.e.}, Softmax loss) usually lacks the power of feature discrimination for deep face recognition. To address this issue, one group proposes to explore the mining-based loss functions \cite{OHEM,Focal,Lift,Yuan}. Shrivastava \textit{et al.} \cite{OHEM} develop a hard mining softmax (HM-Softmax) to improve the feature discrimination by constructing mini-batches using high-loss examples. Among which, the percentage of hard examples is empirically decided and the easy examples are completely discarded. In contrast, Lin \textit{et al.} \cite{Focal} design a relatively soft mining softmax, namely Focal loss (F-Softmax), to focus training on a sparse set of hard examples. It usually achieves more promising results than the simple hard mining softmax. Yuan \textit{et al.} \cite{Yuan} select the hard examples based on model complexity and train an ensemble to model examples of different hard levels. The other group prefers to design margin-based loss functions \cite{SphereFace,AM-Softmax,Arc-Softmax}. This group does not focus on optimizing hard examples but directly increasing the feature margin between different classes. Wen \textit{et al.} \cite{Center} develop a center loss to learn centers for each identity to enhance the intra-class compactness. Wang \textit{et al.} \cite{NormFace} and Ranjan \textit{et al.} \cite{L2Constrain} propose to use a scale parameter to control the temperature of softmax loss, producing higher gradients to the well-separated samples to shrink the intra-class variance. Liu \textit{et al.} \cite{L-softmax,SphereFace}  introduce an angular margin (A-Softmax) between the ground truth class and other classes to encourage the larger inter-class variance. However, it is usually unstable and the optimal parameters are hard to determinate. To enhance the stability of A-Softmax loss, several alternative approaches \cite{AM-Softmax,EM-Softmax,Cosine,Arc-Softmax} have been proposed. Wang \textit{et al.} \cite{AM-Softmax} design an additive margin (AM-Softmax) loss to stabilize the optimization and have achieved promising performance. Deng \textit{et al.} \cite{Arc-Softmax}  develop an additive angular margin (Arc-Softmax) loss, which has a more clear geometric interpretation.

Although these two groups have been well-verified to learn discriminative features for face recognition. The motivation of mining-based losses is to focus on hard examples while margin-based losses are to enlarge the feature margin between different classes. Currently, they develop independently and both of them have their own intrinsic drawbacks. To the mining-based losses, the definition of hard examples is ambiguous and they are often empirically selected. How to semantically decide the hard examples is still an open problem. To the margin-based losses, most of them learn discriminative features by enlarging the feature margin, only from the perspective of ground truth class (\textit{self-motivation}). They usually ignore the discriminative power from the perspective of other non-ground truth classes (\textit{other-motivation}). Moreover, the relation between mining-based and margin-based losses remains unclear.

To overcome the above shortcomings, this paper tries to design a new loss function, which adaptively emphasizes on the informative support vectors to bridge the gap between mining-based and margin-based losses and semantically integrate them into one framework. To sum up, the main contributions of this paper can be summarized as follows:
\begin{itemize}
\item{We propose a novel SV-Softmax loss, which eliminates the ambiguity of hard examples as well as absorbs the discriminative power of other classes by focusing on support vectors. To the best of our knowledge, this is the first attempt to semantically fuse the mining-based and margin-based losses into one framework.}
\item{We deeply analyze the relations of our SV-Softmax loss to the current mining-based and margin-based losses, and further develop an
improved version SV-X-Softmax loss to enhance the feature discrimiantion. Our code will be available at {\url{https://github.com/xiaoboCASIA/SV-X-Softmax}}.}
\item{We conduct extensive experiments on the benchmarks of LFW \cite{LFW}, MegaFace Challenge \cite{megaface_1,megaface_2} and Trillion Pairs Challenge, which have verified the superiority of our new approach over the baseline Softmax loss, the mining-based Softmax losses, the margin-based Softmax losses, and their naive fusions.}
\end{itemize}

\section{Preliminary Knowledge}
\noindent \textbf{Softmax}. Softmax loss is defined as the pipeline combination of the last fully connected layer, the softmax function and the cross-entropy loss. In face recognition, the weights $\bm{w}_k$, (where $ k \in \{1,2,\dots,K\}$ and $K$ is the number of classes) and the feature $\bm{x}$ of the last fully connected layer are usually normalized and the magnitude is replaced as a scale parameter $s$ \cite{NormFace,AM-Softmax,Arc-Softmax}. In consequence, given an input feature vector $\bm{x}$ with its corresponding ground truth label $y$, the softmax loss can be formulated as follows:
\begin{equation}\label{Softmax}
\begin{aligned}
\mathcal{L}_1 = - \log\frac{e^{s\cos(\theta_{\bm{w}_y,\bm{x}})}}{e^{s\cos(\theta_{\bm{w}_y,\bm{x}})}+\sum_{k\ne y}^Ke^{s\cos(\theta_{\bm{w}_k,\bm{x}})}},
\end{aligned}
\end{equation}
where $\cos(\theta_{\bm{w}_k,\bm{x}})=\bm{w}_k^T\bm{x}$ is the cosine similarity and $\theta_{\bm{w}_k,\bm{x}}$ is the angle between $\bm{w}_k$ and $\bm{x}$. As pointed out by a great many studies \cite{L-softmax,SphereFace,AM-Softmax,Arc-Softmax}, the learned features with softmax loss are prone to be separable, rather than to be discriminative for face recognition.

\noindent \textbf{Mining-based Softmax}. Hard example mining is becoming a common practice to effectively train deep CNNs. Its idea is to focus training on the informative examples, thus it usually results in more discriminative features. There are recent works that select hard examples based on loss value \cite{OHEM,Focal} or model complexity \cite{Yuan} to learn discriminative features. Generally, they can be summarized as:
\begin{equation}\label{Mining-Softmax}
\begin{aligned}
\mathcal{L}_2 = -g(p_y) \log\frac{e^{s\cos(\theta_{\bm{w}_y,\bm{x}})}}{e^{s\cos(\theta_{\bm{w}_y,\bm{x}})}+\sum_{k\ne y}^Ke^{s\cos(\theta_{\bm{w}_k,\bm{x}})}},
\end{aligned}
\end{equation}
where $p_y=\frac{e^{s\cos(\theta_{\bm{w}_y,\bm{x}})}}{e^{s\cos(\theta_{\bm{w}_y,\bm{x}})}+\sum_{k\ne y}^Ke^{s\cos(\theta_{\bm{w}_k,\bm{x}})}}$ is the predicted ground truth probability and $g(p_y)$ is an indicator function. Basically, to the soft mining method Focal loss \cite{Focal} (F-Softmax), $g(p_y)=(1-p_y)^\gamma$, $\gamma$ is a modulating factor. To the hard mining method HM-Softmax \cite{OHEM}, $g(p_y)=0$ when the sample is indicated as easy while $g(p_y)=1$ when the sample is hard. However, the definition of hardness is ambiguous and they usually lead to sensitive performance.

\noindent \textbf{Margin-based Softmax}. To directly enhance the feature discrimination, several margin-based softmax loss functions \cite{SphereFace,EM-Softmax,AM-Softmax,Arc-Softmax} have been proposed in recent years. In summary, they can be defined as follows:
\begin{equation}\label{Margin-Softmax}
\begin{aligned}
\mathcal{L}_3 = - \log\frac{e^{sf(m,\theta_{\bm{w}_y,\bm{x}})}}{e^{sf(m,\theta_{\bm{w}_y,\bm{x}})}+\sum_{k\ne y}^Ke^{s\cos(\theta_{\bm{w}_k,\bm{x}})}},
\end{aligned}
\end{equation}
where $f(m,\theta_{\bm{w}_y,\bm{x}})$ is a carefully designed margin function. Basically,
$f(m_1,\theta_{\bm{w}_y,\bm{x}})= \cos(m_1\theta_{\bm{w}_y,\bm{x}})$ is the motivation of A-Softmax loss \cite{SphereFace}, where $m_1\ge1$ and is an integer. $f(m_2,\theta_{\bm{w}_y,\bm{x}})= \cos(\theta_{\bm{w}_y,\bm{x}})-m_2$ with $m_2 >0$ is the AM-Softmax loss \cite{AM-Softmax}. $f(m_3,\theta_{\bm{w}_y,\bm{x}})= \cos(\theta_{\bm{w}_y,\bm{x}}+m_3)$ with $m_3>0$ is the Arc-Softmax loss \cite{Arc-Softmax}. More generally, the margin function can be summarized into a combined version: $f(m,\theta_{\bm{w}_y,\bm{x}})=\cos(m_1\theta_{\bm{w}_y,\bm{x}} + m_3) - m_2$. However, all these methods achieve the feature margin only from the perspective of ground truth class $y$. They are not aware of the importance of other non-ground truth classes.

\section{Problem Formulation} \label{main}
\subsection{Naive Mining-Margin Softmax Loss}
The mining-based loss functions aim to focus on the hard examples while the margin-based loss functions are to enlarge the feature margin between different classes. Therefore, these two branches can seamlessly incorporate into each other. The naive motivation to directly integrate them can be formulated as:
\begin{equation}\label{Mining-Margin-Softmax}
\begin{aligned}
\mathcal{L}_4 = -g(p_y) \log\frac{e^{sf(m,\theta_{\bm{w}_y,\bm{x}})}}{e^{sf(m,\theta_{\bm{w}_y,\bm{x}})}+\sum_{k\ne y}^Ke^{s\cos(\theta_{\bm{w}_k,\bm{x}})}}.
\end{aligned}
\end{equation}
However, this formulation Eq. (\ref{Mining-Margin-Softmax}) only absorbs their own merits. It can not solve their respective shortcomings. Detailedly, it only encourages the feature margin from the perspective of the ground truth class by $f(m,\theta_{\bm{w}_y,\bm{x}})$ (\textit{self-motivation}), ignoring the feature discriminative power of other non-ground truth classes (\textit{other-motivation}). Moreover, the hard examples are still empirically selected by the indicator function $g(p_y)$, without semantic guidance. In other words, the definition of hard examples is ambiguous.

\begin{figure}[t]
\begin{center}
\includegraphics[width=0.9\linewidth]{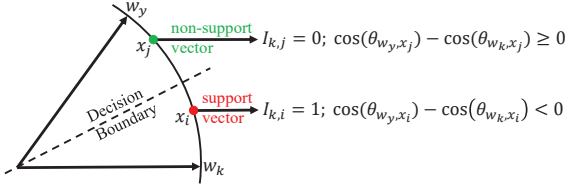}
\end{center}
   \caption{A geometrical interpretation of SV-Softmax loss from feature perspective. The support vectors (red circle points) are those who are mis-classified by the current classifiers. SV-Softmax loss semantically focuses on optimizing such support vectors.}
\label{fig:sv-softmax}
\end{figure}

\subsection{Support Vector Guided Softmax Loss}
Intuition says that considering the well-separated feature vectors has little effect on the learning problem. That means the mis-classified feature vectors are more crucial to enhance the feature discriminability. Motivated by this, the hard example mining \cite{OHEM} and the recent Focal loss \cite{Focal} techniques are proposed to focus training on a sparse set of hard examples and ignore the vast number of easy ones during training. However, they either empirically sample hard examples according to loss values or empirically down-weight the easy examples by a modulating factor. In other words, the definition of hard examples is ambiguous, and without intuitive interpretation.

To address it, we alternatively introduce a more elegant way to focus training on the informative features (\textit{i.e.}, support vectors). Specifically, we define a binary mask to adaptively indicate whether a sample is selected as the support vector by a specific classifier in the current stage. To the end, the binary mask is defined as follows:
\begin{equation}\label{weight}
\begin{aligned}
\ \  I_k =
 \left \{
   \begin{aligned}
  & 0,  \ \ \cos(\theta_{\bm{w}_y,x})-\cos(\theta_{\bm{w}_k,x})\geq 0 \\
  & 1, \ \ \cos(\theta_{\bm{w}_y,x})-\cos(\theta_{\bm{w}_k,x})< 0\\
   \end{aligned}
\right. .
\end{aligned}
\end{equation}
From the definition, we can see that if a sample is mis-classified, \textit{i.e.}, $\cos(\theta_{\bm{w}_y,\bm{x}})-\cos(\theta_{\bm{w}_k,\bm{x}})<0$, it will be emphasized temporarily. In this way, the concept of hard examples is clearly defined and we mainly focus on such a sparse set of support vectors. Consequently, our Support Vector Guided Softmax (\textbf{SV-Softmax}) loss is formulated:
\begin{equation}\label{SV-Softmax}
\mathcal{L}_5 = - \log\frac{e^{s\cos(\theta_{\bm{w}_y,\bm{x}})}}{e^{s\cos(\theta_{\bm{w}_y,\bm{x}})}+\sum_{k\neq y}^K h(t,\theta_{\bm{w}_k,\bm{x}},I_k)e^{s\cos(\theta_{\bm{w}_k,\bm{x}})}},
\end{equation}
where $t$ is a preset hyperparameter and the indicator function $h(t,\theta_{\bm{w}_k,\bm{x}},I_k)$ is defined as:
\begin{equation}
 h(t,\theta_{\bm{w}_k,\bm{x}},I_k)=e^{s(t-1)(\cos(\theta_{\bm{w}_k,\bm{x}})+1)I_k}.
\end{equation}
Obviously, when $t=1$, the designed SV-Softmax loss becomes identical to the original softmax loss. Figure \ref{fig:sv-softmax} gives the geometrical interpretation of our SV-Softmax loss.

\begin{figure}[t]
\begin{center}
 \includegraphics[width=0.495\linewidth]{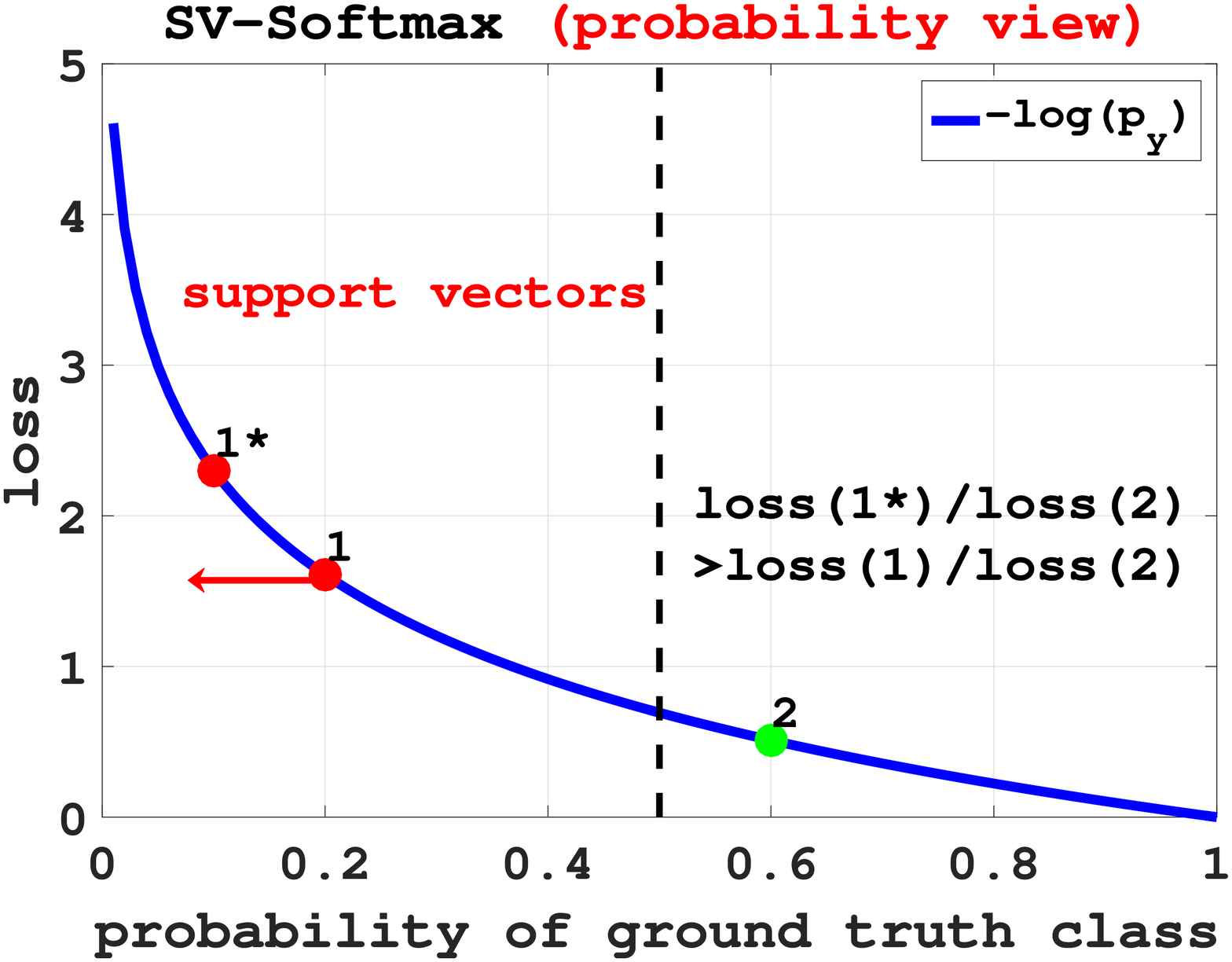}
 \includegraphics[width=0.495\linewidth]{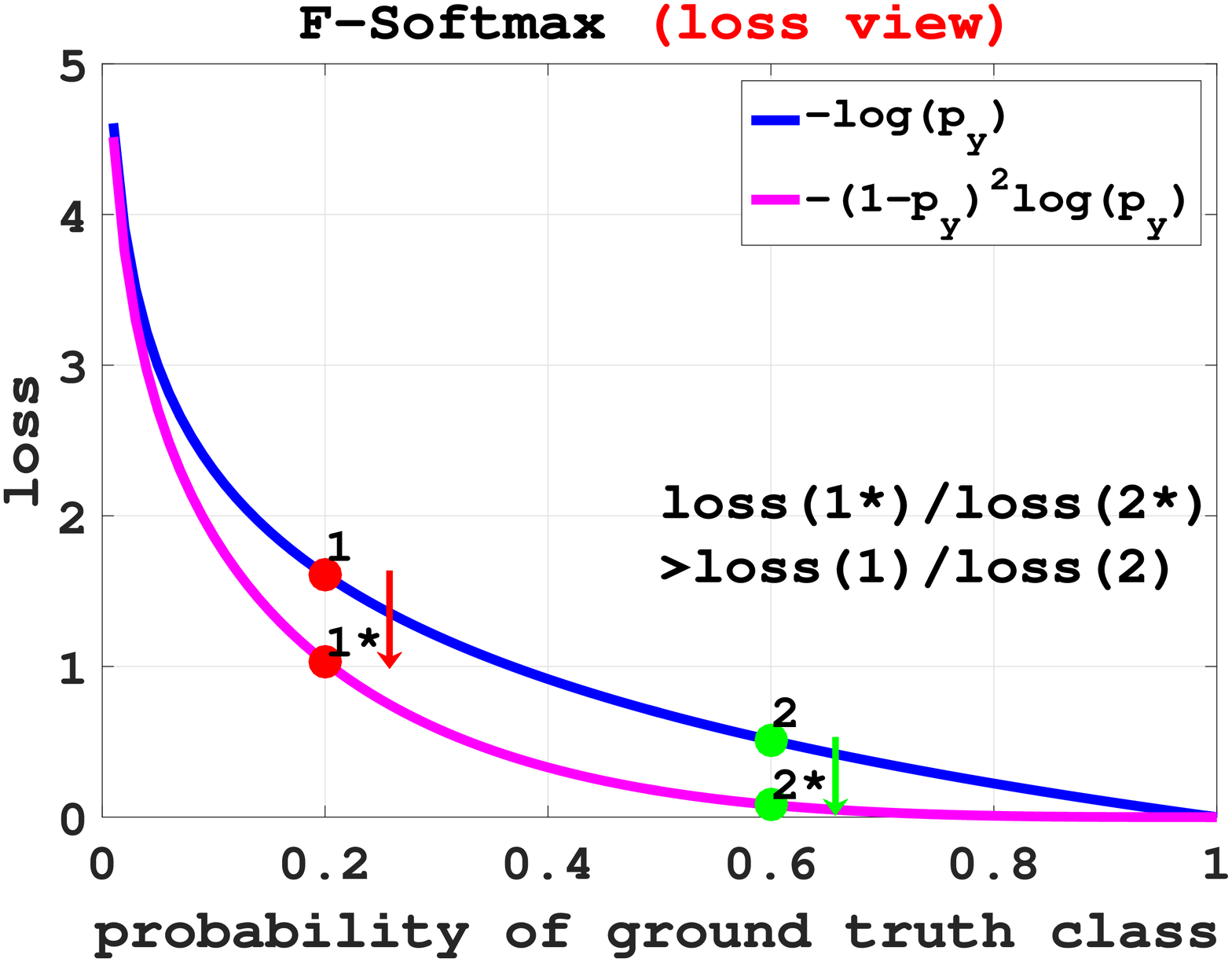}
\end{center}
   \caption{\textbf{From left to right}: SV-Softmax loss vs. Mining-based softmax loss (\textit{e.g.}, Focal loss \cite{Focal}). SV-Softmax loss semantically defines the hard examples (support vectors) and emphasizes them from the probability view, while the hard examples of Focal loss are ambiguous and are concerned from the loss view. }
\label{fig:sv-mining}
\end{figure}

\begin{figure*}[t]
\begin{center}
   \includegraphics[width=0.85\linewidth]{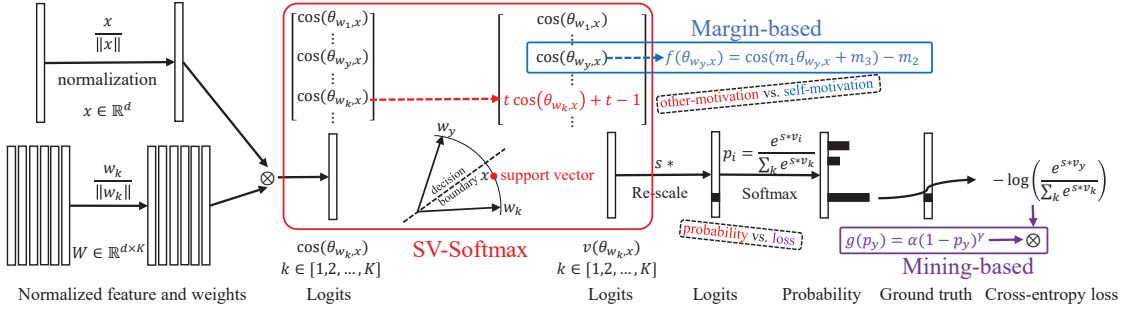}
\end{center}
   \caption{Pipeline of our SV-Softmax loss and its relations to the existing mining-based and margin-based losses. Our SV-Softmax loss semantically integrates the motivation of mining-based and margin-based losses into one framework, but from different viewpoints.}
\label{fig:framework}
\end{figure*}

\begin{figure}[t]
\begin{center}
\includegraphics[width=0.9\linewidth]{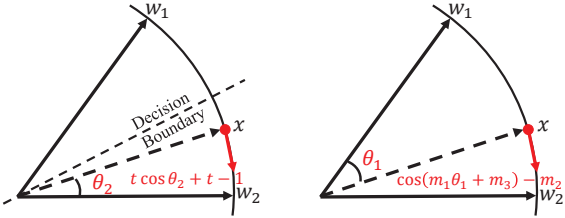}
\end{center}
   \caption{\textbf{From left to right}: SV-Softmax loss vs. Margin-based softmax loss. SV-Softmax loss enlarges the feature margin from other classes (\textit{other-motivation}) while current margin-based losses are directly from the ground truth class (\textit{self-motivation}).}
\label{fig:sv-margin}
\end{figure}

\subsubsection{Relation to Mining-based Softmax Losses}
To illustrate the advantages of our SV-Softmax loss over the traditional mining-based loss functions (\textit{e.g.}, Focal loss \cite{Focal}), we use the binary classification case as an example. Assume that we have two samples $\bm{x}_1$ and $\bm{x}_2$, both of them are from class 1. Figure \ref{fig:sv-mining} gives a diagram, where $\bm{x}_1$ is relatively hard while $\bm{x}_2$ is relatively easy. The traditional mining-based Focal loss is to differentially re-weight the losses of hard and easy examples, such that:
\begin{equation}
 \frac{\rm{loss}_{1^*}}{\rm{loss}_{2^*}} > \frac{\rm{loss}_1}{\rm{loss}_2}.
\end{equation}
In that way, the importance of hard examples is emphasized. This strategy is directly from the loss perspective and the definition of hard examples is ambiguous. While our SV-Softmax loss is from a different way. Firstly, we semantically define the hard examples (support vectors) according to the decision boundary. Then, to the support vector $\bm{x}_1$, we reduce its probability, such that:
\begin{equation}
 \frac{\rm{loss}_{1^*}}{\rm{loss}_2} > \frac{\rm{loss}_1}{\rm{loss}_2}.
\end{equation}
In summary, the differences between SV-Softmax loss and mining-based Focal loss \cite{Focal} are displayed in Figure \ref{fig:sv-mining}.



\subsubsection{Relation to Margin-based Softmax Losses}
Similarly, assume that we have a sample $\bm{x}$ from class 1, and it is a little far way from its ground truth class, (\text{e.g.}, the red circle point in Figure \ref{fig:sv-margin}). The original softmax loss aims to make $\bm{w}_1^T\bm{x}>\bm{w}_2^T\bm{x} \Longleftrightarrow \cos(\theta_1) > \cos(\theta_2)$. To make the objective more rigorous, margin-based losses usually introduce a margin function $f(m,\theta_1)=\cos(m_1\theta_1+m_3)-m_2$ from the perspective of ground truth class \cite{SphereFace,AM-Softmax,Arc-Softmax}:
\begin{equation}
 \cos(\theta_1)\geq f(m,\theta_1)>\cos(\theta_2).
\end{equation}
In contrast, our SV-Softmax loss enlarge the feature margin from the perspective of other non-ground truth classes. Specifically, we have introduced a margin function $h^*(t,\theta_2)$ to these mis-classified features:
\begin{equation}
\cos(\theta_1)>h^*(t,\theta_2)\geq \cos(\theta_2),
\end{equation}
where $h^*(t,\theta_2)=\log[h(t,\theta_2)e^{\cos(\theta_2)}]=t\cos(\theta_2)+t-1$. Our SV-Softmax loss semantically enlarges the feature margin from other non-ground truth classes while margin-based losses make theirs efforts from the ground truth class. For multi-class case, Our SV-Softmax loss is class-specific margins. Figure \ref{fig:sv-margin} gives their geometrical comparison. To sum up, Figure \ref{fig:framework} shows the pipeline of our SV-Softmax loss and its relations to the mining-based and margin-based losses.

\begin{figure}[t]
\begin{center}
\includegraphics[width=0.9\linewidth]{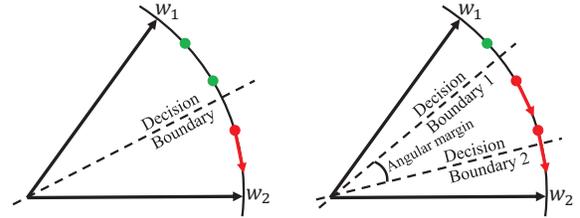}
\end{center}
   \caption{\textbf{From left to right}: SV-Softmax loss vs. SV-X-Softmax loss. To increase the mining range, we adopt the margin-based decision boundaries to select support vectors. Thus the non-support vectors in SV-Softmax may be support vectors in SV-X-Softmax.}
\label{fig:sv-svx}
\end{figure}

\subsubsection{SV-X-Softmax}
According to the above discussions, our SV-Softmax loss semantically fuses the motivation of mining-based and margin-based losses into one framework, but from different viewpoints. Therefore, we can also absorb their strengths into our SV-Softmax loss. Specifically, to increase the mining range, we adopt the margin-based decision boundaries to indicate the support vectors. Consequently, the improved SV-X-Softmax loss can be formulated as:
\begin{equation}\label{SV-X-Softmax}
\noindent \mathcal{L}_6 = -\log  \frac{e^{sf(m,\theta_{\bm{w}_y,\bm{x}})}}{e^{sf(m,\theta_{\bm{w}_y,\bm{x}})}+\sum_{k\neq y}^K h(t,\theta_{\bm{w}_k,\bm{x}},I_k)e^{s\cos(\theta_{\bm{w}_k,\bm{x}})}},
\end{equation}
where X is the margin-based losses. It can be A-Softmax \cite{SphereFace}, AM-Softmax \cite{AM-Softmax} and Arc-Softmax \cite{Arc-Softmax} \textit{etc}. The indicator mask $I_k$ is re-computed according to margin-based decision boundaries\footnote{That why we uniformity call the hard examples as "support vectors", because it is similar to the definition in \cite{cortes1995support}.}. Specifically,
\begin{equation}\label{weightX}
\begin{aligned}
\ \  {I}_k =
 \left \{
   \begin{aligned}
  & 0,  \ \ f(m,\theta_{\bm{w}_y,x})-\cos(\theta_{\bm{w}_k,x})\geq 0 \\
  & 1, \ \ f(m,\theta_{\bm{w}_y,x})-\cos(\theta_{\bm{w}_k,x})< 0\\
   \end{aligned}
\right. .
\end{aligned}
\end{equation}
Figure \ref{fig:sv-svx} gives the geometrical illustration of our SV-X-Softmax loss. It is best because from the motivation of margin-based losses, SV-X-Softmax loss enlarges the feature margin by integrating the self-motivation of ground truth class and the other-motivation of other classes into one framework. While from the motivation of mining-based losses, it semantically enlarges the mining range.

\section{Optimazation}
In this section, we show that the proposed SV-Softmax loss (\ref{SV-Softmax}) is trainable and can be easily optimized by the typical stochastic gradient descent. The difference between the original softmax loss and the proposed SV-Softmax loss lies in the last fully connected layer
$\bm{v}=[v_1,v_2,\dots,v_K]^T=[\cos(\theta_{\bm{w}_1,\bm{x}}),\cos(\theta_{\bm{w}_2,\bm{x}}),\dots,\cos(\theta_{\bm{w}_K,\bm{x}})]^T$.


To the forward, when $k=y$, it is the same as the original softmax loss (\textit{i.e.}, $v_y=\cos(\theta_{\bm{w}_y,\bm{x}})$). When $k\neq y$, it has two cases, if the feature vector is easy for an specific class, it is the same as the original softmax (\textit{i.e.},  $v_k=\cos(\theta_{\bm{w}_k,\bm{x}})$). Otherwise, it will be recomputed as $\log[h(t,\theta_{\bm{w}_k,\bm{x}})e^{\cos(\theta_{\bm{w}_k,\bm{x}})}]=t\cos(\theta_{\bm{w}_k,\bm{x}})+t-1$. To the backward propagation, we use the chain rule to compute the partial derivative. The derivative of $\bm{W}$ and the CNN feature $\bm{x}$ of the last fully connected layer should be re-emphasized:
\begin{equation} \label{UpdateW}
\frac{\partial{\mathcal{L}_5}}{\partial{\bm{W}}} =
\left\{
\begin{aligned}
\frac{\partial{\mathcal{L}_5}}{\partial{\bm{v}}}\frac{\partial{\bm{v}}}{\partial{\bm{w}_y}} &= \frac{\partial{\mathcal{L}_5}}{\partial{\bm{v}}}\bm{x}, && k = y \\
\frac{\partial{\mathcal{L}_5}}{\partial{\bm{v}}}\frac{\partial{\bm{v}}}{\partial{\bm{w}_k}} &= \frac{\partial{\mathcal{L}_5}}{\partial{\bm{v}}}\bm{x}, &&k \neq y; v_y \geq v_k \\
\frac{\partial{\mathcal{L}_5}}{\partial{\bm{v}}}\frac{\partial{\bm{v}}}{\partial{\bm{w}_k}} &=t \frac{\partial{\mathcal{L}_5}}{\partial{\bm{v}}}\bm{x}, && k \neq y; v_y < v_k
\end{aligned}
\right.
\end{equation}

\begin{equation} \label{Updatex}
\frac{\partial{\mathcal{L}_5}}{\partial{\bm{x}}} =
\left\{
\begin{aligned}
\frac{\partial{\mathcal{L}_5}}{\partial{\bm{v}}}\frac{\partial{\bm{v}}}{\partial{\bm{x}}} &= \frac{\partial{\mathcal{L}_5}}{\partial{\bm{v}}}\bm{w}_y, && k=y \\
\frac{\partial{\mathcal{L}_5}}{\partial{\bm{v}}}\frac{\partial{\bm{v}}}{\partial{\bm{x}}} &= \frac{\partial{\mathcal{L}_5}}{\partial{\bm{v}}}\bm{w}_k, && k\neq y;  v_y \geq v_k \\
\frac{\partial{\mathcal{L}_5}}{\partial{\bm{v}}}\frac{\partial{\bm{v}}}{\partial{\bm{x}}} &=t \frac{\partial{\mathcal{L}_5}}{\partial{\bm{v}}}\bm{w}_k, && k\neq y; v_y < v_k
\end{aligned}
\right.
\end{equation}
where the computation form of $\frac{\partial{\mathcal{L}_5}}{\partial{\bm{v}}}$ is the same as the original softmax loss.
The whole scheme for a single image is summarized in Algorithm \ref{algorithm}. It is trivial to perform derivation with mini-batch input. Moreover, it is also straightforward to the SV-X-Softmax loss case.

\begin{algorithm}[t] \label{algorithm}
	\SetAlgoLined \caption{\small SV-Softmax}
	\begin{small}
		\KwIn{A CNN feature $\bm{x}$ with its corresponding label $y$. Initialized parameters $\bm{\Theta}$ in convolution layers. Parameter $\bm{W}$ in the last fully connected layer. The learning rate $\lambda$ and the indicator parameter $t$. The number of iteration $\alpha \leftarrow 0$.}
		
		\While{not converged}{

          1: $\alpha\leftarrow \alpha+1$;

          2: According to the definition of hard examples (\ref{weight}), we compute the SV-Softmax loss by (\ref{SV-Softmax});

          3: Compute the back-propagation error of each CNN feature $\bm{x}$ by (\ref{Updatex}) and the weight $\bm{W}$ by (\ref{UpdateW});

          4: Update the parameters $\bm{W}$ and $\bm{\Theta}$ by $\bm{W}^{(\alpha+1)} = \bm{W}^{(\alpha)} - \lambda^{(\alpha)} \frac{\partial{\mathcal{L}_5}}{\partial{\bm{W}^{(\alpha)}}}$; \\
          \ \ \ $\bm{\Theta}^{(\alpha+1)}=\bm{\Theta}^{(\alpha)}-\lambda^{(\alpha)}\frac{\partial{\mathcal{L}_5}}{\partial{\bm{x}^{(\alpha)}}} \frac{\partial{\bm{x}^{(\alpha)}}}{\partial{\bm{\Theta}^{(\alpha)}}}$;

		}
		\KwOut{Parameters $\bm{\Theta}$ and $\bm{W}$.}
	\end{small}
\end{algorithm}

\begin{table*}
\begin{center}
\begin{tabular}{|c|c|c|c|c|c| }
\hline
& \multirow{2}{*}{Method}  & LFW 6000 & LFW BLUFR   & LFW BLUFR   & LFW BLUFR    \\
&         & Pairs Accuracy & TPR@FAR=1e-3      & TPR@FAR=1e-4      & TPR@FAR=1e-5       \\

\hline\hline
Baseline & Softmax                       & 99.26 & 99.46 &	98.44 & 95.24 \\
\hline
\multirow{2}{*}{Mining-based}& F-Softmax  \cite{Focal}       & 99.46 & 99.62 & 98.76 & 95.97 \\
& HM-Softmax \cite{OHEM}        & 99.26 & 99.48 & 98.48 & 95.11 \\
\hline
\multirow{3}{*}{Margin-based} & A-Softmax  \cite{SphereFace}  & 99.36 & 99.68 & 99.09 & 97.20 \\
& Arc-Softmax\cite{Arc-Softmax} & 99.63 & 99.86 & 99.68 & 98.18 \\
& AM-Softmax \cite{AM-Softmax}  & 99.61 & 99.86 & 99.75 & 98.18 \\
\hline
\multirow{4}{*}{Naive-fused}& F-Arc-Softmax & 99.66	& 99.87 & 99.73 & 98.32 \\
& F-AM-Softmax  & 99.66 & 99.87 & 99.76 & 98.39 \\
& HM-Arc-Softmax                & 99.51	& 99.86 & 99.70	& 98.74	 \\
& HM-AM-Softmax                 & 99.63 & 99.87 & 99.75 & 98.90 \\
\hline
\multirow{3}{*}{Ours}& SV-Softmax   & 99.48 & 99.78 & 99.39 & 98.14  \\
& SV-Arc-Softmax                    & \textbf{99.78} & 99.85  & 99.77 & 98.52 \\
& SV-AM-Softmax                     & 99.76 & \textbf{99.87}  & \textbf{99.81} & \textbf{99.22} \\
\hline
\end{tabular}
\end{center}		
\caption{Verification performance (\%) of different loss functions on LFW test data.}
\label{LFW}
\end{table*}

\section{Experiments}

\subsection{Datasets}

\noindent \textbf{Training Data}. The MS-Celeb-1M dataset \cite{Msceleb} contains about 100k identities with 10 million images. However, it consists of a great many noisy face images. Fortunately, the trillionpairs consortium has made their efforts to get a high-quality version MS-Celeb-1M-v1c, which is well-cleaned with 86,876 identities and 3,923,399 aligned images.


\noindent \textbf{Validation Data}. We employ Labelled Faces in the Wild (LFW) \cite{LFW} as the validation data. LFW contains 13,233 web-collected images from 5,749 different identities, with large variations in pose, expression and illuminations.


\noindent \textbf{Test Data}. We use two datasets, MegaFace \cite{megaface_1} and Trillion Pairs\footnote{\url{http://trillionpairs.deepglint.com/overview}}, as the test data. MegaFace datasets aim at evaluating the performance of face recognition algorithms at the million scale of distractors, which include gallery set and probe set. The gallery set, a subset of Flickr photos from Yahoo, consists of more than one million images from 690,000 different individuals. The probe set has two existing databases: Facescrub \cite{facescrub} and FGNET \cite{fgnet}. In this study, we use the Facescrub as the probe set, which contains 100,000 photos of 530 unique individuals, wherein 55,742 images are males, and 52,076 images are females. Trillion Pairs datasets are recently released as a public available testing benchmark, which are consisted of the following two parts, ELFW and DELFW. ELFW is the face images of celebrities in LFW name list. There are 274,000 images from 5,700 identities. DELFW is the distractors for ELFW. There are in total 1.58 million face images from Flickr.

\subsection{Experimental Settings}
\noindent \textbf{Data Processing}.
We detect the faces by adopting the FaceBoxes detector \cite{facebox} and localize five landmarks (two eyes, nose tip and two mouth corners) through a simple 6-layer CNN \cite{feng2017wing}. The detected faces are cropped and resized to 120$\times$120, and each pixel (ranged between [0,255]) in RGB images is normalized by subtracting 127.5 and then being divided by 128. For all the training faces, they are horizontally flipped with probability 0.5 for data augmentation.

\noindent \textbf{CNN Architecture}. In face recognition, there are many kinds network architectures \cite{SphereFace,AM-Softmax,wang2018devil}. To be fair, the CNN architecture should be the same to test different loss functions. As suggested by the work \cite{wang2018devil}, we use Attention-56 \cite{Attention56} as our baseline architecture to achieve a good balance between computation and accuracy. The output of Attention-56 has and finally gets a 512-dimension feature by the operation of averaging pooling. The scale parameter $s$ has already been discussed sufficiently in previous works \cite{AM-Softmax,cosface}. In this paper, we directly fixed it to 30. For details, the adopted Attention-56 architecture is provided in supplementary materials.

%


\noindent \textbf{Training}. All the CNN models are trained with stochastic gradient descent (SGD) algorithm and trained from scratch, with the batch size of 32 on 4 P40 GPUs parallelly, total batch size 128. The weight decay is set to 0.0005 and the momentum is 0.9. The learning rate is initially 0.1 and divided by 10 at the 100k, 160k, 220k iterations, and we finish the training process at 240k iterations.

\noindent \textbf{Test}. At the testing stage, only the features of original image are employed (512-dimension) to compose the face representation. All the reported results in this paper are evaluated by a single model, without model ensemble or other fusion strategies.

To the evaluation metrics, the cosine distance of features is computed as the similarity score. Face identification and verification are conducted by ranking and thresholding the scores. Specifically, for face identification, the Cumulative Match Characteristics (CMC) curves are adopted to evaluate the Rank-1 face identification accuracy. For face verification, the Receiver Operating Characteristic (ROC) curves are adopted. The true positive rate (TPR) at low false acceptance rate (FAR) is emphasized since in real applications false acceptance gives higher risks than false rejection. We test our models on several popular public face datasets, including LFW \cite{LFW}, MegaFace Challenge \cite{megaface_1,megaface_2} and the recent Trillion Pairs Challenge. Specifically, for LFW, the unrestricted with labeled outside data on 6000 pairs accuracy \cite{LFW} and the BLUFR \cite{blufr} protocols are reported. For Megaface Challenge, the identification Rank-1 accuracy and the verification rate TPR@FAR =1e-6 are reported. For Trillion Pairs Challenge, every pair between ELFW and DELFW is used. There are in total 0.4 trillion pairs. To the face identification task, they provide a 1.58 million-size gallery and a 270k-size query for top-1 identification and the metric TPR@FAR=1e-3 is reported. While to the face verification task, the verification rate TPR@FAR=1e-9 is reported. For more details about the protocols, please refer to the works \cite{LFW,blufr,megaface_1}.

To the compared methods, we compare our method with the baseline Softmax loss (\textbf{Softmax}) and the recently proposed state-of-the-arts, including 2 mining-based softmax losses (\textit{i.e.}, hard example mining (\textbf{HM-Softmax} \cite{OHEM}) and Focal loss (\textbf{F-Softmax} \cite{Focal})), 3 margin-based softmax losses (the angular Softmax loss (\textbf{A-Softmax}\cite{Angular}), the additive margin Softmax loss (\textbf{AM-Softmax}\cite{AM-Softmax}), and the additive angular margin Softmax loss (\textbf{Arc-Softmax}\cite{Arc-Softmax})) and their 4 naive fusions (\textbf{F-AM-Softmax}, \textbf{F-Arc-Softmax}, \textbf{HM-AM-Softmax} and \textbf{HM-Arc-Softmax}). For all the compared methods, their source codes can be downloaded from the github or from authors' webpages. The corresponding parameters are determined according to their suggestions (\textit{e.g.}, the feature margin parameter $m$ is 0.35 for AM-Softmax and is 0.5 for Arc-Softmax). For more details, please refer to the supplementary materials.

\begin{figure}[t]
\begin{center}
   \includegraphics[width=0.495\linewidth]{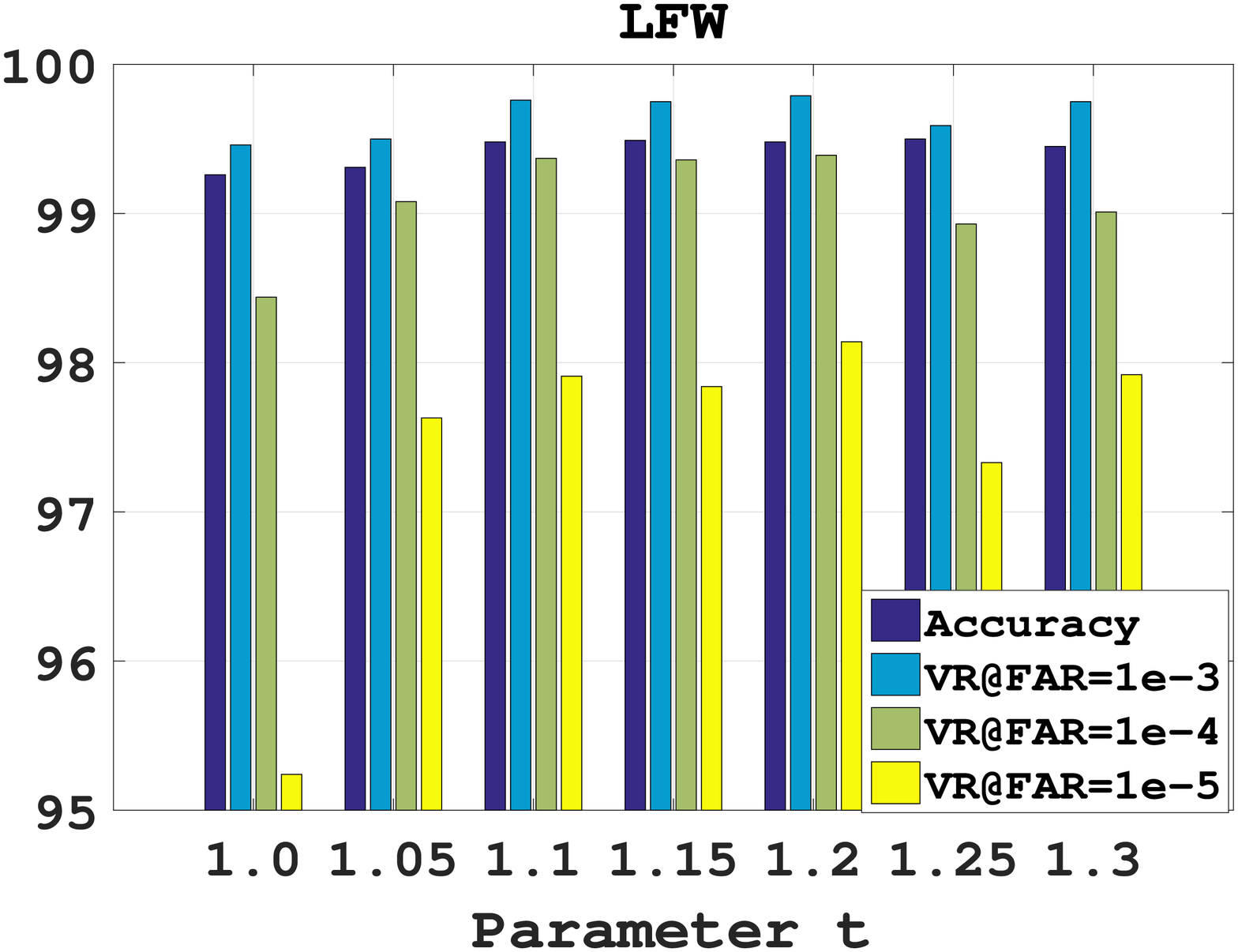}
   \includegraphics[width=0.495\linewidth]{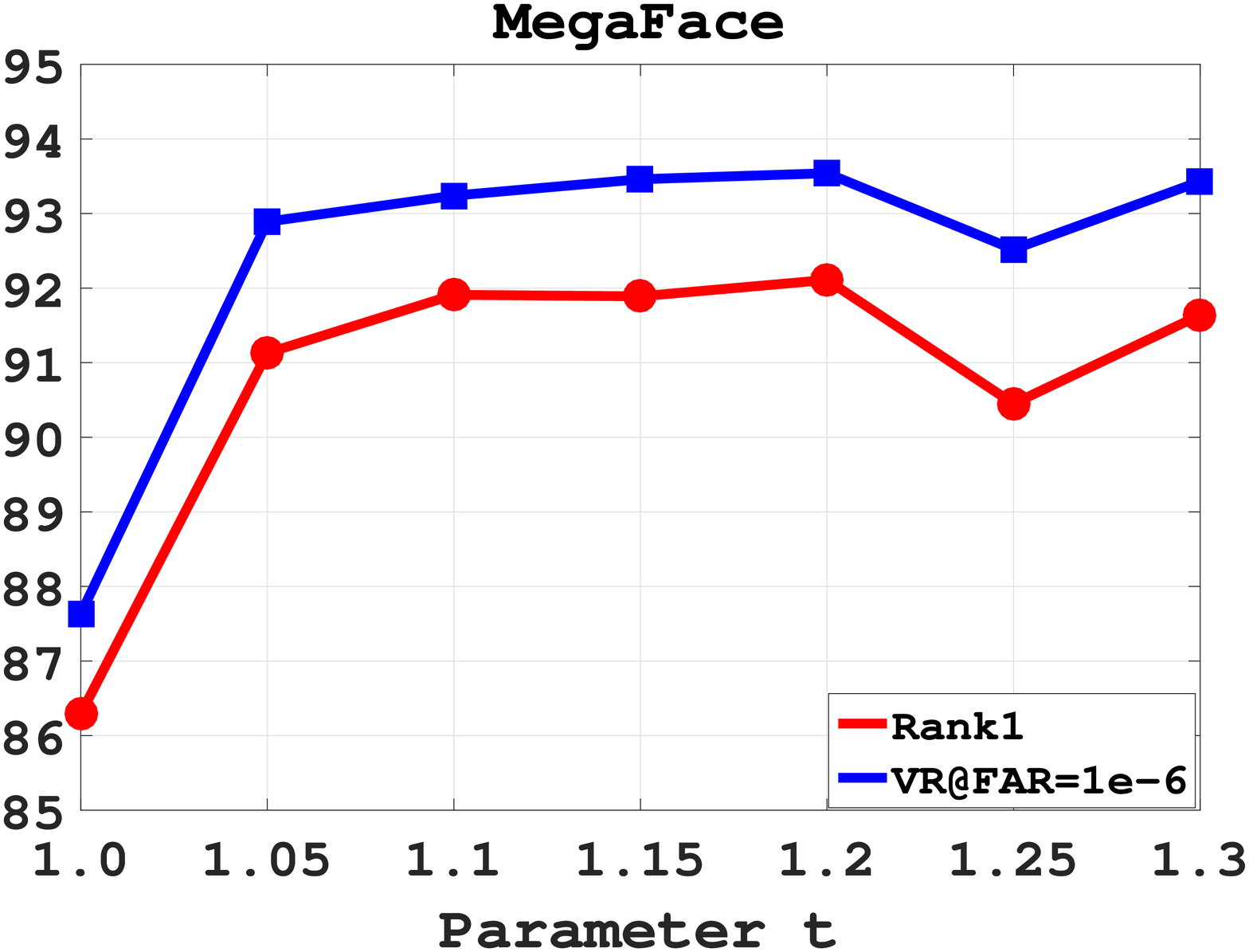}
\end{center}
   \caption{\textbf{From left to right}: Identification and Verification performance (\%) of SV-Softmax loss with different indicator parameter $t$ on LFW and MegaFace, respectively. }
\label{fig:parameter}
\end{figure}

\subsection{Effects of indicator parameter $t$}
Since the indicator parameter $t$ plays an important role in the developed SV-Softmax loss, we first conduct experiments to
search its possible best value. By varying t from 1.0 to 1.3 (If t is larger than 1.4, the model may fail to converge), we use the Attention-56 network and the SV-Softmax loss to train models on the MS-Celeb-1M-v1c dataset and evaluate its performance on the validation set LFW. As illustrated in the left sub-figure of Figure \ref{fig:parameter}, with $t$ being increased, the 6000 pairs accuracy and the BLUFR of LFW are improved consistently, and get saturated at $t=1.2$. This demonstrates the effectiveness of our SV-Softmax loss (compared $t\neq 1.0$ with $t=1.0$). To validate the sensitivity of our indicator parameter $t$, we directly use the trained models to test them on MegaFace, the effects are reported in the right sub-figure of Figure \ref{fig:parameter}. From the curves, we can see that our SV-Softmax loss is insensitive to the indicator parameter $t$ in a certain range. According to this study, $t$ is set to fixed 1.2 in the subsequent experiments.

\begin{table}
\begin{center}
\begin{tabular}{|c|c|c| }
\hline
\multirow{2}{*}{Method}  & Identification     & Verification  \\
        & Rank1@1e6          & TPR@FAR=1e-6 \\
\hline\hline
Softmax                         & 86.29 & 87.63  \\
\hline
F-Softmax   \cite{Focal}        & 88.29 & 89.83 \\
HM-Softmax  \cite{OHEM}         & 86.58 & 88.39 \\
\hline
A-Softmax   \cite{SphereFace}   & 88.54 & 89.40  \\
Arc-Softmax \cite{Arc-Softmax}  & 93.67	& 94.47  \\
AM-Softmax  \cite{AM-Softmax}   & 94.77	& 95.44  \\
\hline
F-Arc-Softmax                   & 93.98 & 95.10 \\
F-AM-Softmax                    & 94.47 & 94.84 \\
HM-Arc-Softmax                  & 94.05 & 95.26 \\
HM-AM-Softmax                   & 94.78 & 95.57 \\
\hline
SV-Softmax             & 92.11 & 93.54   \\
SV-Arc-Softmax         & 97.14 & \textbf{97.57}  \\
SV-AM-Softmax          & \textbf{97.20} & 97.38  \\
\hline
\end{tabular}
\end{center}		
\caption{Results (\%) of different losses on MegaFace Challenge.}
\label{MegaFace}
\end{table}

\subsection{Experiments on LFW}
Table \ref{LFW} provides the quantitative results of all the competitors on LFW dataset. The bold number in each column represents the best performance. To the 6000 pairs accuracy protocol, it is well-known that this protocol is typical and easy for deep face recognition, and all the competitors can achieve over 99\% accuracy rate. So the improvement of our SV-Softmax loss is not quite large. From the numbers, we observe that the naive fusions of mining-based and margin-based losses, \textit{e.g.}, HM-AM-Softmax and F-AM-Softmax, outperform the simple mining-based or margin-based ones.  Despite this, our imporved SV-AM-Softmax still achieves about 0.3\% improvements. To the BLUFR protocol, the similar trends as the 6000 pairs accuracy, our improved SV-AM-Softmax loss achieves the best performance among all the competitors. Due to the evaluation protocols on LFW are nearly to be saturated, it would be better to test our models on MegaFace and Trillion Pair Challenges.

\begin{figure}[t]
\begin{center}
   \includegraphics[width=0.495\linewidth]{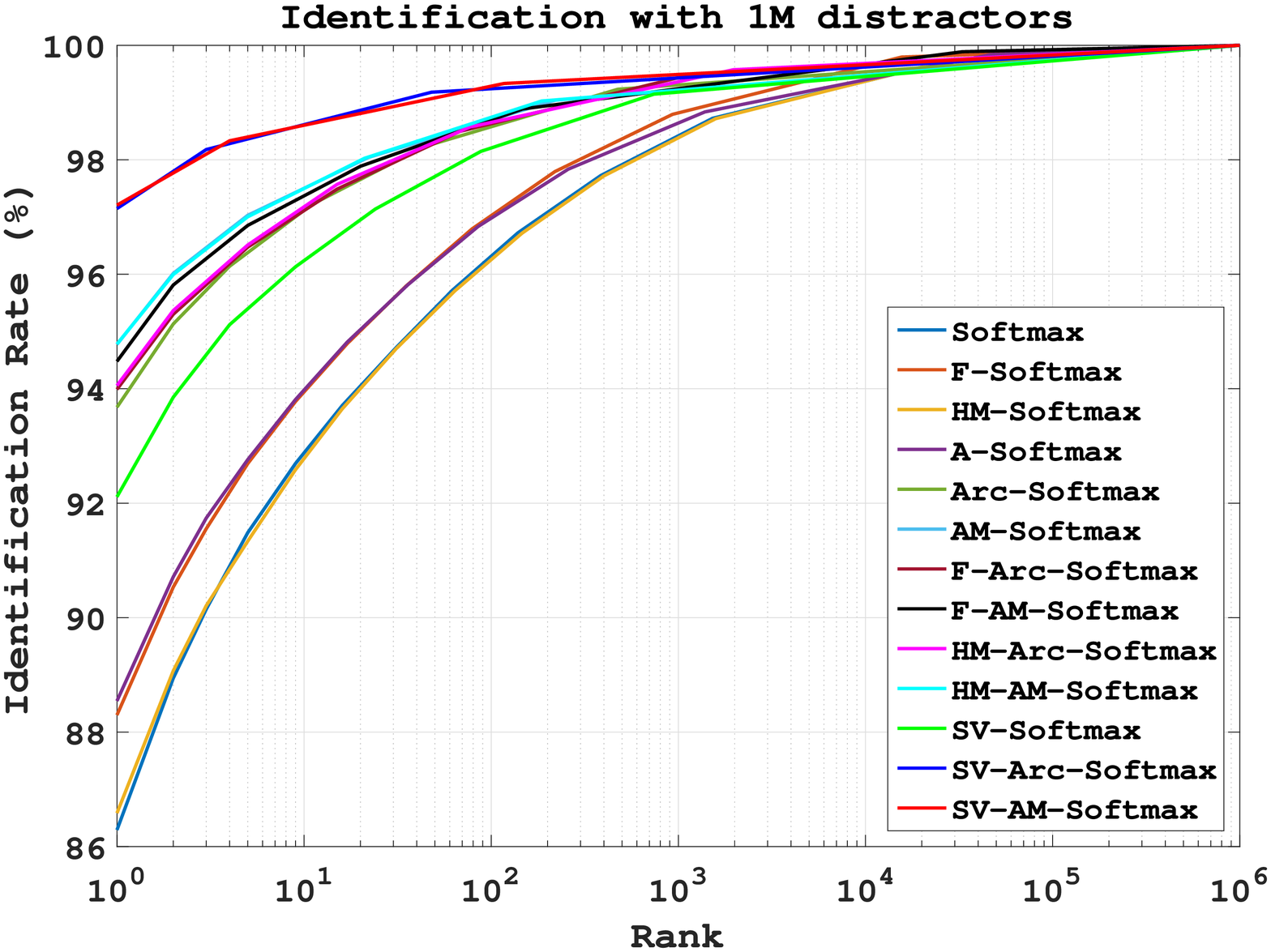}
   \includegraphics[width=0.495\linewidth]{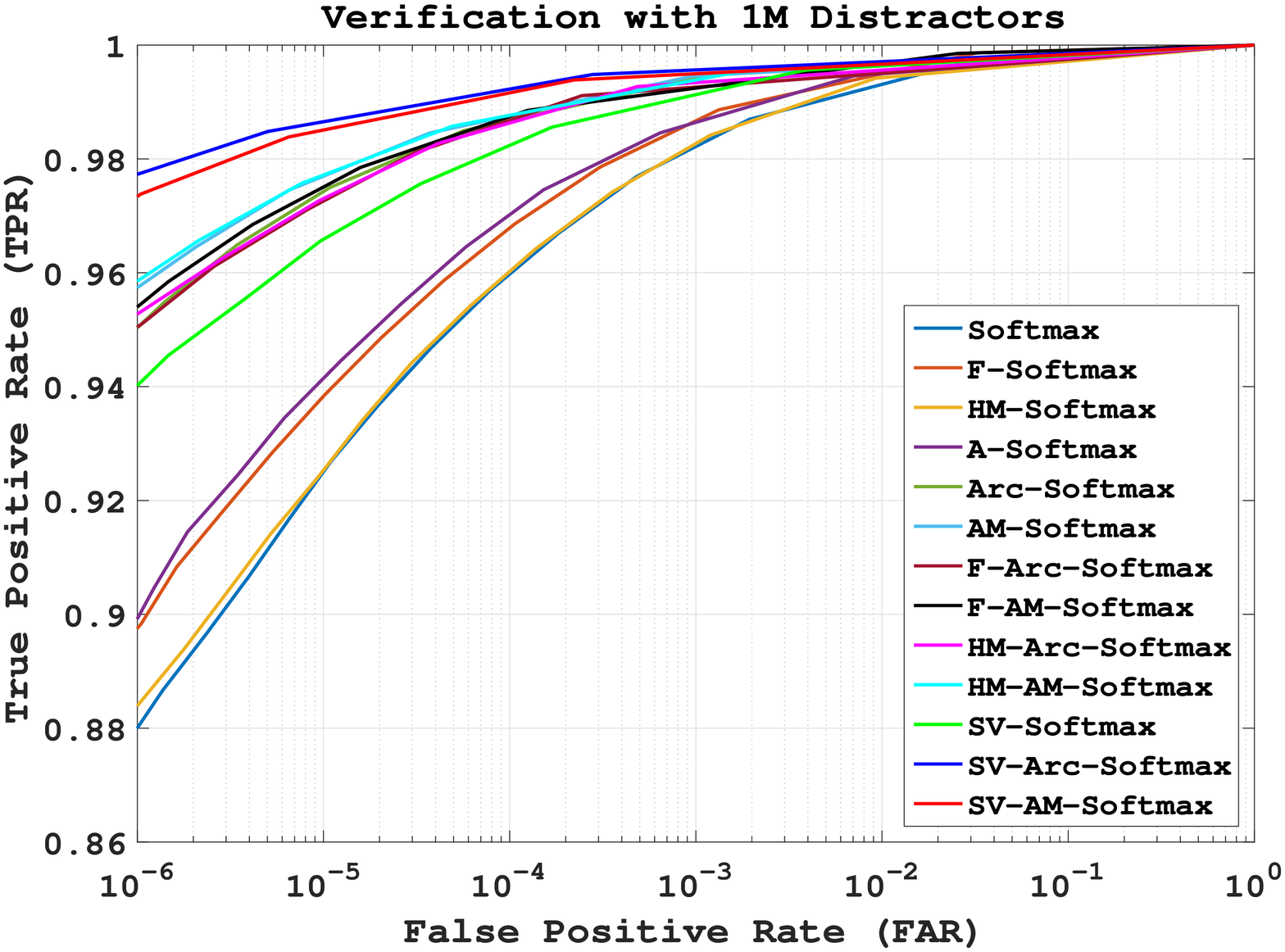}
\end{center}
   \caption{\textbf{Left}: CMC curves of different loss functions with 1M distractors on MegaFace \cite{megaface_1} Set 1. \textbf{Right}: ROC curves of different loss functions with 1M distractors on MegaFace \cite{megaface_1} Set 1.}
\label{fig:mega}
\end{figure}

\subsection{Experiments on MegaFace Challenge}
Table \ref{MegaFace} shows the identification and verification results on MegaFace dataset. In particular, compared with the baseline Softmax loss and the mining-based Softmax losses, our SV-Softmax loss achieves at least 3\% improvements at both the Rank-1 identification rate and the verification TPR@FAR=1e-6 rate. The reason is that our SV-Softmax loss has clearly defined the hard examples (\textit{i.e.}, support vectors), thus it is better than existing mining-based losses. While compared with the margin-based Softmax losses, the performance of our SV-Softmax loss is slightly lower than them. This is reasonable because the support vectors decided by the Softmax decision boundary in SV-Softmax loss may not be enough for learning discriminative features. Our improved versions SV-Arc-Softmax and SV-AM-Softmax losses, wherein the support vectors are determined by the margin-based decision boundaries, can further boost the performance because they absorb the complementary merits of margin-based losses. Specifically, to our SV-AM-Softmax loss, it beats the best margin-based competitor AM-Softmax loss by a large margin (about 2.4\% at Rank-1 identification rate and 1.9\% verification rate). Compared with the naive fusions of mining-based and margin-based losses, our improved SV-AM-Softmax loss is also better than them. It is about 2.4\% higher at Rank-1 identification rate and 1.8\% higher at verification rate than the second best competitor HM-AM-Softmax loss. To sum up, our imporved SV-X-Softmax losses, which eliminate the ambiguity of hard examples as well as absorb the discriminative power of other classes by focusing on support vectors, are inherently the best in the current stage. In Figure \ref{fig:mega}, we draw both of the CMC curves to evaluate the performance of face identification and the ROC curves to evaluate the performance of face verification on MegaFace Set 1. From the curves, we can see the similar trends at other measures. In this experiment, our SV-Softmax loss with its improved version SV-AM-Softmax approach have shown their superiority for both the identification and verification tasks.

\begin{table}
\begin{center}
\scalebox{0.96}{
\begin{tabular}{|c|c|c| }
\hline
\multirow{2}{*}{Method}  & Identification     & Verification  \\
        & TPR@FAR=1e-3  & TPR@FAR=1e-9 \\
\hline\hline
Softmax                         & 36.61 & 33.87  \\
\hline
F-Softmax   \cite{Focal}        & 39.80 & 37.14 \\
HM-Softmax  \cite{OHEM}         & 36.75 & 34.46 \\
\hline
A-Softmax   \cite{SphereFace}   & 43.89 & 43.76  \\
Arc-Softmax \cite{Arc-Softmax}  & 57.48	& 57.45  \\
AM-Softmax  \cite{AM-Softmax}   & 61.80	& 61.61  \\
\hline
F-Arc-Softmax                   & 56.80 & 56.87 \\
F-AM-Softmax                    & 61.85 & 61.79 \\
HM-Arc-Softmax                  & 55.93 & 56.63 \\
HM-AM-Softmax                   & 61.42 & 61.33 \\
\hline
SV-Softmax             & 51.18 & 46.78   \\
SV-Arc-Softmax         & 71.19 & 70.33  \\
SV-AM-Softmax          & \textbf{73.56} & \textbf{72.71}  \\
\hline
\end{tabular}}
\end{center}		
\caption{Performance (\%) of different loss functions on Trillion Pairs Challenge. }
\label{deeplint}
\end{table}

\subsection{Experiments on Trillion Pairs Challenge}
Table \ref{deeplint} displays the performance comparison on the recent Trillion Pairs Challenge, from which, we can conclude that the results exhibit the same trends that emerged on LFW and MegaFace datasets. Besides, the trends are more obvious. Concretely, both of the current mining-based and margin-based losses are better than the simple softmax loss for face recognition. However, the margin-based losses usually achieve higher performance than the mining-based losses, because the motivation of margin-based losses is to enhance the feature discrimination while the motivation of mining-based losses is to focus training on hard examples. Their naive fusions can slightly improve the performance further. However, the naive fusions are still suffering from the ambiguity of hard examples and the lack of discriminative power of other classes. Therefore, they are limited for face recognition. Our SV-X-Softmax (\textit{e.g.}, SV-AM-Softmax) losses absorb the strengths and discard the drawbacks of the current ming-based and margin-based loss functions, thus they achieve the highest performance.

\begin{table}
\begin{center}
\scalebox{0.9}{
\begin{tabular}{|c|c|c|c|}
\hline
 LFW       & LFW BLUFR & LFW BLUFR & LFW BLUFR  \\
6000 & TPR  & TPR  & TPR \\
Accuracy & @FAR=1e-3 & @FAR=1e-4 & @FAR=1e-5\\
\hline\hline
99.85 (our) & 99.92 &	99.89 & 99.13 \\
\textbf{99.87} (1st) & - - & - - & - - \\
\hline
\end{tabular}}
\end{center}		
\caption{Performance (\%) of SV-AM-Softmax loss on LFW.}
\label{LFW}
\end{table}

\begin{table}
\begin{center}
\scalebox{0.9}{
\begin{tabular}{|c|c|}
\hline
 MegaFace Identification & MegaFace Verification   \\
 Rank-1@1e6   & TPR@FAR=1e-6  \\
\hline\hline
98.82 (our) & 99.03 (our) \\
\textbf{99.93} (1st) & \textbf{99.93} (1st) \\
\hline
\end{tabular}}
\end{center}		
\caption{Performance (\%) of SV-AM-Softmax loss on MegaFace.}
\label{MegaFace}
\end{table}

\begin{table}
\begin{center}
\scalebox{0.9}{
\begin{tabular}{|c|c|}
\hline
 Trillion Pairs Identification & Trillion Pairs Verification   \\
 TPR@FAR=1e-3       & TPR@FAR=1e-9  \\
\hline\hline
82.25 (our) & 78.49 (our) \\
\textbf{85.67} (1st) & \textbf{82.29} (1st) \\
\hline
\end{tabular}}
\end{center}		
\caption{Performance (\%) of SV-AM-Softmax loss on Trillion
Pairs.}
\label{Trillion}
\end{table}


\section{Improvement by Designing Architectures}

To further boost the performance, we try to make the adopted Attention-56 \cite{Attention56} architecture deeper. Specifically, we change the stages of [1,1,1] used in Attention-56 into [3,6,2]. Moreover, inspired by \cite{Arc-Softmax}, we incorporate the IRSE module into the architecture. The results are displayed in Tables \ref{LFW}-\ref{Trillion}. Note that all current results are training based on the simple MS-Celeb-1Mv1c dataset and only the single model performance is reported. From the numbers, we can see that our SV-AM-Softmax loss has achieved the competitive absolute performance. In the future, it would be better to fuse the MS1M-ArcFace \cite{Arc-Softmax} and Asian datasets\footnote{\url{http://trillionpairs.deepglint.com/data}} and design model ensemble methods (\textit{e.g.}, feature concatenation).


\section{Conclusion}
This paper has proposed a simple but very effective loss function, namely support vector guided softmax loss (\textit{i.e.}, SV-Softmax), for face recognition. In specific, SV-Softmax loss explicitly concentrates on optimizing the support vectors. Thus it semantically integrates the motivation of mining-based and margin-based loss functions into one framework. Consequently, it is intrinsically better than the current mining-based losses, margin-based losses and their naive fusions. Extensive experiments on several benchmark datasets have clearly demonstrated the advantages of our new approach over the state-of-the-art alternatives.

{\small
\bibliographystyle{ieee}
\bibliography{egbib_sv}

\begin{thebibliography}{10}\itemsep=-1pt

\bibitem{fgnet}
Fg-net aging database.
\newblock \url{http://www.fgnet.rsunit.com/}.
\newblock 2010.

\bibitem{cai2014support}
S.~Cai, W.~Zuo, L.~Zhang, X.~Feng, and P.~Wang.
\newblock Support vector guided dictionary learning.
\newblock In {\em ECCV}, 2014.

\bibitem{cortes1995support}
C.~Cortes and V.~Vapnik.
\newblock Support-vector networks.
\newblock {\em Machine learning}, 20(3), 1995.

\bibitem{Arc-Softmax}
J.~Deng, J.~Guo, and S.~Zafeiriou.
\newblock Arcface: Additive angular margin loss for deep face recognition.
\newblock {\em arXiv preprint arXiv:1801.07698}, 2018.

\bibitem{feng2017wing}
Z.-H. Feng, J.~Kittler, M.~Awais, P.~Huber, and X.-J. Wu.
\newblock Wing loss for robust facial landmark localisation with convolutional
  neural networks.
\newblock {\em arXiv preprint arXiv:1711.06753}, 2017.

\bibitem{Msceleb}
Y.~Guo, L.~Zhang, Y.~Hu, X.~He, and J.~Gao.
\newblock Ms-celeb-1m: A dataset and benchmark for large-scale face
  recognition.
\newblock In {\em ECCV}, 2016.

\bibitem{Resnet}
K.~He, X.~Zhang, and S.~Ren.
\newblock Deep residual learning for image recognition.
\newblock In {\em CVPR}, 2016.

\bibitem{LFW}
G.~Huang, M.~Ramesh, T.~Berg, and E.~Miller.
\newblock Labeled faces in the wild: A database for studying face recognition
  in unconstrained enviroments.
\newblock In {\em Technical Report}, 2007.

\bibitem{megaface_1}
I.~Kemelmacher-Shlizerman, S.~M. Seitz, D.~Miller, and E.~Brossard.
\newblock The megaface benchmark: 1 million faces for recognition at scale.
\newblock In {\em CVPR}, 2016.

\bibitem{SM-Softmax}
X.~Liang, X.~Wang, Z.~Lei, S.~Liao, and S.~Li.
\newblock Soft-margin softmax for deep classification.
\newblock In {\em ICONIP}, 2017.

\bibitem{blufr}
S.~Liao, Z.~Lei, D.~Yi, and S.~Z. Li.
\newblock A benchmark study of large-scale unconstrained face recognition.
\newblock In {\em ICB}, 2014.

\bibitem{Focal}
Y.~Lin, P.~Goyal, and R.~Girshick.
\newblock Focal loss for dense object detection.
\newblock In {\em ICCV}, 2017.

\bibitem{L-softmax}
W.~Liu, Y.~Wen, and Z.~Yu.
\newblock Large-margin softmax loss for convolutional neural networks.
\newblock In {\em ICML}, 2016.

\bibitem{SphereFace}
W.~Liu, Y.~Wen, Z.~Yu, M.~Li, and L.~Song.
\newblock Sphereface: Deep hypersphere embedding for face recognition.
\newblock In {\em CVPR}, 2017.

\bibitem{Cosine}
Y.~Liu, H.~Li, and X.~Wang.
\newblock Learning deep features via congenerous cosine loss for person
  recognition.
\newblock In {\em ICCV}, 2017.

\bibitem{megaface_2}
A.~Nech and I.~Kemelmacher-Shlizerman.
\newblock Level playing field for million scale face recognition.
\newblock In {\em CVPR}, 2017.

\bibitem{facescrub}
H.-W. Ng and S.~Winkler.
\newblock A data-driven approach to cleaning large face datasets.
\newblock In {\em ICIP}, 2014.

\bibitem{DeePFR}
O.~Parkhi, A.~Vedaldi, and A.~Zisserman.
\newblock Deep face recognition.
\newblock In {\em BMVC}, 2015.

\bibitem{L2Constrain}
R.~Ranjan, C.~Castillo, and R.~Chellappa.
\newblock L2-constrained softmax loss for discriminative face verification.
\newblock {\em arXiv preprint arXiv:1703.09507.}, 2017.

\bibitem{Facenet}
F.~Schroff, D.~Kalenichenko, and J.~Philbin.
\newblock Facenet: A unified embedding for face recognition and clustering.
\newblock In {\em CVPR}, 2015.

\bibitem{shi2017cross}
H.~Shi, X.~Wang, D.~Yi, Z.~Lei, X.~Zhu, and S.~Z. Li.
\newblock Cross-modality face recognition via heterogeneous joint bayesian.
\newblock {\em IEEE Signal Processing Letters}, 24(1):81--85, 2017.

\bibitem{OHEM}
A.~Shrivastava, A.~Gupta, and R.~Girshick.
\newblock Training region-based object detectors with online hard example
  mining.
\newblock In {\em CVPR}, 2016.

\bibitem{VGG}
K.~Simonyan and Z.~Andrew.
\newblock Very deep convolutional networks for large-scale image recognition.
\newblock {\em arXiv preprint arXiv:1409.1556}, 2014.

\bibitem{Lift}
H.~Song, Y.~Xiang, S.~Jegelka, and S.~Savarese.
\newblock Deep metric learning via lifted structured feature embedding.
\newblock In {\em CVPR}, 2016.

\bibitem{DeepID}
Y.~Sun, Y.~Chen, and X.~Wang.
\newblock Deep learning face representation by joint
  identification-verification.
\newblock In {\em NIPS}, 2014.

\bibitem{Contrastive}
Y.~Sun, X.~Wang, and X.~Tang.
\newblock Deep learning face representation from predicting 10,000 classes.
\newblock In {\em CVPR}, 2014.

\bibitem{DeepID2+}
Y.~Sun, X.~Wang, and X.~Tang.
\newblock Deeply learned face representations are sparse, selective, and
  robust.
\newblock In {\em CVPR}, 2015.

\bibitem{Deepface}
Y.~Taigman, M.~Yang, and M.~Ranzato.
\newblock Deepface: Closing the gap to human-level performance in face
  verification.
\newblock In {\em CVPR}, 2014.

\bibitem{wang2018devil}
F.~Wang, L.~Chen, C.~Li, S.~Huang, Y.~Chen, C.~Qian, and C.~C. Loy.
\newblock The devil of face recognition is in the noise.
\newblock In {\em ECCV}, 2018.

\bibitem{AM-Softmax}
F.~Wang, J.~Cheng, W.~Liu, and H.~Liu.
\newblock Additive margin softmax for face verification.
\newblock {\em IEEE Signal Processing Letters}, 25(7):926--930, 2018.

\bibitem{Attention56}
F.~Wang, M.~Jiang, C.~Qian, S.~Yang, C.~Li, H.~Zhang, X.~Wang, and X.~Tang.
\newblock Residual attention network for image classification.
\newblock {\em arXiv preprint arXiv:1704.06904}, 2017.

\bibitem{NormFace}
F.~Wang, X.~Xiang, J.~Chen, and A.~Yuille.
\newblock Normface: $ l_2 $ hypersphere embedding for face verification..
\newblock In {\em ACM MM}, 2017.

\bibitem{cosface}
H.~Wang, Y.~Wang, Z.~Zhou, X.~Ji, Z.~Li, D.~Gong, J.~Zhou, and W.~Liu.
\newblock Cosface: Large margin cosine loss for deep face recognition.
\newblock {\em arXiv preprint arXiv:1801.09414}, 2018.

\bibitem{Angular}
J.~Wang, F.~Zhou, and S.~Wen.
\newblock Deep metric learning with angular loss.
\newblock In {\em ICCV}, 2017.

\bibitem{USSDL}
X.~Wang, X.~Guo, and S.~Z. Li.
\newblock Adaptively unified semi-supervised dictionary learning with active
  points.
\newblock In {\em ICCV}, 2015.

\bibitem{EM-Softmax}
X.~Wang, S.~Zhang, Z.~Lei, S.~Liu, X.~Guo, and S.~Z. Li.
\newblock Ensemble soft-margin softmax loss for image classification.
\newblock {\em arXiv preprint arXiv:1805.03922}, 2018.

\bibitem{Center}
Y.~Wen, K.~Zhang, and Z.~Li.
\newblock A discriminative feature learning approach for deep face recognition.
\newblock In {\em ECCV}, 2016.

\bibitem{wright2009robust}
J.~Wright, A.~Y. Yang, A.~Ganesh, S.~S. Sastry, and Y.~Ma.
\newblock Robust face recognition via sparse representation.
\newblock {\em PAMI}, 2009.

\bibitem{Yuan}
Y.~Yuan, K.~Yang, and C.~Zhang.
\newblock Hard-aware deeply cascaded embedding.
\newblock In {\em ICCV}, 2017.

\bibitem{facebox}
S.~Zhang, X.~Zhu, Z.~Lei, H.~Shi, X.~Wang, and S.~Z. Li.
\newblock Faceboxes: A cpu real-time face detector with high accuracy.
\newblock In {\em IJCB}, 2017.

\bibitem{zheng2018ring}
Y.~Zheng, D.~K. Pal, and M.~Savvides.
\newblock Ring loss: Convex feature normalization for face recognition.
\newblock In {\em CVPR}, 2018.

\end{thebibliography}
}

\end{document}